\documentclass[letterpaper, 10 pt, conference]{ieeeconf}  

\IEEEoverridecommandlockouts                              

\usepackage{graphics} 
\usepackage{epsfig} 
\usepackage{times} 
\usepackage{amsmath} 
\usepackage{amssymb}  
\let\proof\relax 
\let\endproof\relax
\usepackage{amsthm}
\usepackage[T1]{fontenc}
\usepackage{soul}
\usepackage{url}
\usepackage[hidelinks]{hyperref}
\usepackage[utf8]{inputenc}
\usepackage[small]{caption}
\usepackage{booktabs}
\usepackage{algorithm}
\usepackage[noend]{algpseudocode}
\usepackage{algorithmicx}
\usepackage[switch]{lineno}
\usepackage{subcaption}
\usepackage{cleveref}

\newtheorem{theorem}{Theorem}
\newtheorem{lemma}{Lemma}
\newenvironment{proofsketch}{%
  \proof}{\endproof}

\usepackage[dvipsnames]{xcolor}

\title{\LARGE \bf
Solving Multi-Agent Target Assignment and Path Finding with a Single Constraint Tree
}

\author{Yimin Tang$^1$, Zhongqiang Ren$^1$, Jiaoyang Li$^1$ and Katia Sycara$^1$
\thanks{$^{1}$Robotics Institute,
        Carnegie Mellon University, 5000 Forbes Ave, Pittsburgh, PA 15213, USA.
        {\tt\small \{yimint, zhongqir, jiaoyanl, sycara\}@andrew.cmu.edu}}%
}

\begin{document}

\maketitle
\thispagestyle{empty}
\pagestyle{empty}

\begin{abstract}

The Combined Target-Assignment and Path-Finding (TAPF) problem requires simultaneously assigning targets to agents and planning collision-free 
paths for them from their start locations to their assigned targets.
As a leading approach to addressing TAPF, Conflict-Based Search with Target Assignment (CBS-TA) leverages K-best target assignments to create multiple search trees and Conflict-Based Search (CBS) to resolve collisions in each tree.
While CBS-TA finds optimal solutions, it faces scalability challenges due to the duplicated collision resolution in multiple trees and the expensive computation of K-best assignments.
We introduce Incremental Target Assignment CBS (ITA-CBS) to bypass these two computational bottlenecks.
ITA-CBS generates only a single search tree and avoids computing K-best assignments by incrementally computing new 1-best assignments during the search.
We show that ITA-CBS, in theory, is guaranteed to find optimal solutions and, in practice, runs faster than CBS-TA in 96.1\% of 6,334 test cases.
\end{abstract}

\section{Introduction}

The Multi-Agent Path Finding (MAPF) problem requires planning collision-free paths for multiple agents from their respective start locations to pre-assigned target locations while minimizing the sum of path costs~\cite{standley2010finding}.
Solving MAPF to optimality is NP-hard~\cite{yu2013structure}, and many algorithms have been developed to handle this computational challenge.
Among them, Conflict-Based Search (CBS)~\cite{sharon2015conflict} is a widely used approach that finds optimal solutions to MAPF.

This work considers a variant of MAPF that is often referred to as Combined Target-Assignment and Path-Finding (TAPF)~\cite{10.5555/2936924.2937092,honig2018conflict}, where the target locations of the agents are not pre-assigned but need to be allocated during the computation: TAPF requires assigning each agent a unique target (location) out of a pre-specified set of candidate targets and then finds collision-free paths for the agents so that the sum of path costs is minimized.
When the candidate target set of each agent contains only a single target, TAPF becomes MAPF and is thus NP-hard.

MAPF and TAPF arise in many applications such as robotics~\cite{ho2019multi}, computer gaming~\cite{hagelback2015hybrid}, warehouse automation~\cite{li2021lifelong}, traffic management at road intersections~\cite{dresner2008multiagent}.
Several attempts~\cite{honig2018conflict, ma2016multi} have been made to solve TAPF optimally by leveraging MAPF algorithms such as CBS~\cite{sharon2015conflict}.
Among them, a leading approach is Conflict-Based Search with Target Assignment (CBS-TA)~\cite{honig2018conflict}, which simultaneously explores different target assignments and creates multiple search trees (i.e., a CBS forest) while planning collision-free paths with respect to each assignment.

CBS-TA suffers from poor scalability as the number of agents or targets increases for the following two reasons.
First, CBS-TA may resolve the same collision in multiple search trees many times, leading to duplicated computation and low search efficiency.
Second, CBS-TA involves solving a K-best target assignment~\cite{chegireddy1987algorithms,murty1968algorithm} problem, which is often computationally expensive.
We thus attempt to bypass these two computational bottlenecks by exploring a new framework for integrating CBS with target assignment. The resulting algorithm is called Incremental Target Assignment CBS (ITA-CBS).
First, ITA-CBS creates only a single search tree, thereby avoiding duplicated collision resolution in different trees, as seen in CBS-TA.
Second, ITA-CBS eliminates the need to solve the K-best assignment problem. Instead, it updates the target assignment in an incremental manner during the CBS-like search, which further reduces the computational effort. 
Our experimental results show significant improvement in efficiency:
ITA-CBS is faster than CBS-TA in 96.1\% of the test cases, 5 times faster in 38.7\% of the test cases, and 100 times faster in 5.6\% of the test cases, as evaluated across 6,334 test cases.


\section{Problem Definition}

We define the Combined Target-Assignment and Path-Finding (TAPF) problem as follows.
Let $I=\{1,2,\cdots,N\}$ denote a set of $N$ agents.
Let $G = (V,E)$ denote an undirected graph, where each vertex $v \in V$ represents a possible location of an agent in the workspace, and each edge $e \in E$ is a unit-length edge between two vertices that moves an agent from one vertex to the other. Self-loop edges are allowed, which represent ``wait-in-place'' actions.
Each agent $i\in I$ has a unique start location $s_i \in V$.
Let $\{g_j \in V | j \in \{1, 2, ..., M\}\}$, $M\geq N$, denote the set of $M$ target locations.
Let $A$ denote a binary $N \times M$ \emph{target matrix}, where each entry $A[i][j]$ (the $i$-th row and $j$-th column in $A$) is one if agent $i$ is eligible to be assigned to target $g_j$ and zero otherwise. For convenience, we refer to the set of target locations $\{g_j\}$ with $A[i][j]=1$ as the \emph{target set} for agent $i$. 
Our task is to assign each agent $i$ a unique target $g_j$ from its target set and plan corresponding collision-free paths.

Each action of agents, either waiting in place or moving to an adjacent vertex, takes a time unit.
Let $p^i=[v_0^{i}, v_1^{i}, ..., v_{T^{i}}^{i}]$ denote a path of agent $i$ from $v_0^{i}$ to $v_{T^{i}}^{i}$, 
where $v^i_t \in V$ denotes the location of agent $i$ at timestep $t$. We assume that agents rest at their targets after completing their paths, i.e., $v_t^i = v_{T^{i}}^{i}, \forall t \ge T^i$. We consider two types of agent-agent conflicts (i.e., collisions) along their paths.
The first type is the \emph{vertex conflict}, where two agents $i,j$ occupy the same vertex at the same timestep. 
The second type is the \emph{edge conflict}, where two agents go through the same edge from opposite directions at the same timestep.
We use $(i, j, t)$ to denote a vertex/edge conflict between agents $i$ and $j$ at timestep $t$.
It is important to note that the requirement of being conflict-free implies that the target locations assigned to the agents must be distinct from each other.

The goal of the TAPF problem is to find a set of paths $\{p^i | i\in I\}$ for all agents such that, for each agent $i$:
\begin{enumerate}
\item $v_0^{i} = s_i$ (i.e., agent $i$ starts from its start location);
\item $v_{t}^{i} = g_j, \forall t \ge T^{i}$ and $A[i][j] = 1$ (i.e., agent $i$ stops at a target location $g_j$ in its target set);
\item Every pair of adjacent vertices in path $p^i$ is either identical or connected by an edge (i.e., $v_{t}^{i}=v_{t+1}^{i} \lor (v_{t}^{i}, v_{t+1}^{i}) \in E, \forall t \ge 0$);
\item $\{p^i | i\in I\}$ is conflict-free; and
\item The \emph{flowtime} $\sum_{i=1}^{N}T^{i}$ is minimized.
\end{enumerate}

\section{Related Work}
\subsection{MAPF}

MAPF can be viewed as a special case of TAPF where the size of the target set for each agent is one.
MAPF has a long history~\cite{geramifard2006biased,silver2005cooperative} and remains an active research problem~\cite{varambally2022mapf,andreychuk2022multi}. 
A variety of methods are developed to address MAPF, trading off completeness and optimality for runtime efficiency.
These methods range from decoupled methods~\cite{silver2005cooperative,luna2011push,wang2008fast}, which plan a path for each agent independently and synthesize the paths, to coupled methods~\cite{standley2010finding}, which plan for all agents together.
Among them, Conflict-Based Search (CBS)~\cite{sharon2015conflict} is a leading (centralized) optimal MAPF algorithm and forms the foundation of this paper.

CBS is a two-level search algorithm. 
Its low level plans a shortest path for an agent from its start location to its target location. Its high level searches a binary Constraint Tree (CT). 
Each CT node \(H = (c, \Omega, \pi)\) includes a constraint set \(\Omega\), a plan \(\pi\), which is a set of shortest paths for all agents from their start locations to their target locations that satisfy \(\Omega\), and a cost \(c\), which is the flowtime of \(\pi\). 
When expanding \(H\), CBS selects and resolves the first conflict in \(H.\pi\), even when multiple conflicts occur in \(H.\pi\). It formulates two constraints, wherein each constraint prohibits one agent from executing its originally intended action at the conflicting timestep, and adds them to two successor nodes, respectively.
We define two types of constraints, namely vertex constraint $(i, v, t)$ that prohibits agent $i$ from occupying vertex $v$ at timestep $t$ and edge constraint $(i, u, v, t)$ that prohibits agent $i$ from going from vertex $u$ to vertex $v$ at timestep $t$. 
By maintaining a priority queue based on the cost of each node, CBS is provably optimal with respect to the flowtime minimization. 

\subsection{Assignment Problem and TAPF}

Given $N$ agents, $M$ tasks, and a $N\times M$ matrix denoting the corresponding assignment cost of each task to each agent, the task assignment problem~\cite{du1998handbook,munkres1957algorithms,kuhn1955hungarian} seeks to allocate the tasks to agents such that each agent is assigned to a unique task and the total assignment cost is minimized.
Popular methods used to address this problem include the Hungarian algorithm~\cite{munkres1957algorithms,kuhn1955hungarian} and the Successive Shortest Path (SSP) algorithm~\cite{ford1956maximal,busacker1960procedure}.
Additionally, the Dynamic Hungarian algorithm~\cite{mills2007dynamic} aims to quickly re-compute an optimal assignment based on the existing assignment when some entries change in the cost matrix.

TAPF can be viewed as a combination of the MAPF problem and the target assignment problem. 
While MAPF has a pre-defined target for each agent, TAPF involves simultaneously assigning targets to agents and finding conflict-free paths for them.
The leading algorithms for solving TAPF optimally include CBM~\cite{10.5555/2936924.2937092}, which combines CBS with maxflow algorithms to minimize makespan (i.e., $max\{T^i\}$), and CBS-TA~\cite{honig2018conflict}, which construct a CBS forest to minimize flowtime. Our work is built upon CBS-TA.

CBS-TA operates on the following principle: a fixed Target Assignment (TA) solution transforms a TAPF problem into a MAPF problem, and each MAPF problem corresponds to a CT. CBS-TA efficiently explores all nodes of various CTs (CBS forest) by enumerating every TA solution.
Each CT node \(H = (c, \Omega, \pi, \pi_{ta}, r)\) in CBS-TA has two extra fields compared to that in CBS: a TA solution \(\pi_{ta}\), that assigns each agent a unique target location, and a root flag \(r\) signifying if \(H\) is a root. Two nodes have the same TA solution if and only if they belong to the same CT.
CBS-TA maintains a priority queue to store the nodes from all CTs and lazily generates roots with different TA solutions for different CTs.
Because the cost of a root equals the total assignment cost of its TA solution, CBS-TA will not expand a root if there is another root in the priority queue with a TA solution of lower total assignment cost. 
Consequently, CBS-TA first generates only one root node with the optimal TA solution. It then generates a new root with the succeeding optimal TA solution only when the current one has been expanded.
Motivated by K-best task assignment algorithms~\cite{chegireddy1987algorithms,murty1968algorithm} and SSP with Dijkstra algorithm, CBS-TA finds the succeeding optimal TA solution with a time complexity of $O(N^2M^2)$.

Many TAPF variants have been extensively explored. For instance, researchers have extended TAPF to scenarios where each agent can be assigned multiple targets, requiring them to visit these targets sequentially~\cite{HenkelAROS19,nguyen2019generalized,ren21ms}. It's noteworthy that, owing to the success of CBS-TA, numerous extensions~\cite{ren23cbss,10.5555/2936924.2937092,chen2021integrated,ZhongICRA22,okumura2023solving} follow a similar CBS forest approach. Therefore, although our primary focus in this paper is on classic TAPF, our proposed algorithm has the potential to accelerate these extension works as well.

\begin{algorithm}[th]
\small
\caption{ITA-CBS algorithm}
\label{alg:ITA-CBS1}
\textbf{Input}: Graph \(G\), start locations $\{s_i\}$, target locations $\{g_i\}$, target matrix $A$\\
\textbf{Output}: Optimal TAPF solution

\begin{algorithmic}[1] 
\State OPEN = PriorityQueue()
\State \(\Omega^0\) = $\emptyset$
\For{\textbf{each} $(i,j) \in \{1, \cdots, N\} \times \{1, \cdots, M\}$}
\If {$A[i][j]=1$}
    \State \(M_c^0\)[$i$][$j$] = shortestPathSearch($G$, $s_i$, $g_j$, \(\Omega_0\))
\Else
    \State \(M_c^0\)[$i$][$j$] = $\infty$
\EndIf
\EndFor
\State \(\pi_{ta}^0\) = optimalTargetAssignment(\(M_c^0\))
\State \(c^0, \pi^0\) = getPlan(\(\pi_{ta}^0\), \(M_c^0\))
\State \(H_{0}\) = \{\(c^0, \Omega^0, \pi^0, \pi_{ta}^0, M_c^0\)\}
\State Insert \(H_{0}\) to OPEN
\While{OPEN not empty}
\State \(H_{cur}\) = OPEN front node; OPEN.pop()
\State Validate \(H_{cur}.\pi\) until a conflict occurs
\If {\(H_{cur}.\pi\) has no conflict}
    \State \textbf{return} \(H_{cur}.\pi\)
\EndIf
\State ($i, j, t$) = getFirstConflict($H_{cur}.\pi$)
\For{\textbf{each} agent $k$ in ($i,j$)}
\State $Q$ = \(H_{cur}\)
\If{($i, j, t$) is vertex conflict}
\State \(Q.\Omega\) = \(Q.\Omega\) $\cup$ ($k$, $v^k_{t}$, $t$) 
\Else 
\State \(Q.\Omega\) = \(Q.\Omega\) $\cup$ ($k$, $v^k_{t-1}$, $v^k_{t}$, $t$) 
\EndIf
\For{\textbf{each} $x$ with $A[k][x]=1$}
    \State \(Q.M_c\)[$k$][$x$] = shortestPathSearch($G$, $s_k$, $g_x$, \(Q.\Omega\))
\EndFor
\State \(Q.\pi_{ta}\) = optimalTargetAssignment(\(Q.M_c\))
\State \(Q.c, Q.\pi\) = getPlan(\(Q.\pi_{ta}\), \(Q.M_c\))
\State Insert $Q$ to OPEN
\EndFor
\EndWhile
\State \textbf{return} No valid solution
\end{algorithmic}
\end{algorithm}

\section{ITA-CBS}

Our ITA-CBS has the same low-level search as CBS and CBS-TA but a different high-level search. 
Each CT node \(H = (c, \Omega, \pi, \pi_{ta}, M_c)\) in ITA-CBS has two extra fields compared to that in CBS: a TA solution \(\pi_{ta}\) and a $N\times M$ cost matrix $M_c$. Each entry $M_c[i][j]$ of $M_c$ is the cost of the shortest path from $s_i$ to $g_j$ that satisfies the constraint set \(\Omega\)\footnote{In our implementation, we also store this shortest path so that, after we determine \(\pi_{ta}\), we can construct $\pi$ directly from these stored paths.} if $A[i][j]=1$ (i.e., target $g_{j}$ is included in the target set of agent $i$) and $\infty$ otherwise. $\pi_{ta}$ is the optimal TA solution based on $M_c$. $\pi$ is the set of the shortest paths for all agents with respect to $\pi_{ta}$ that satisfies $\Omega$. $c$ is the flowtime of $\pi$, which is identical to the total assignment cost of $\pi_{ta}$. 

As shown in \Cref{alg:ITA-CBS1}, ITA-CBS begins by creating the root node with an empty \(\Omega\) and the corresponding $M_c$ and $\pi_{ta}$ (Lines 2-10).
It maintains a priority queue to store all CT nodes that are generated during the search (Lines 1, 11-13, 28).
In each iteration, ITA-CBS selects a node \(H_{cur}\) with the minimum cost from the priority queue and checks if its plan is conflict-free. If so, this plan is guaranteed to be an optimal solution (Lines 13-16).
Otherwise, ITA-CBS uses the first detected conflict (Line 17) to create two new constraints as in CBS.
It then creates two child nodes identical to \(H_{cur}\) and adds each constraint respectively to the constraint set of the two child nodes (Lines 18-23).
For each new node \(Q\) (with a constraint on agent $k$ added), the low-level search is invoked for agent $k$ to recompute the optimal paths from its start location to all possible targets subject to the new constraint set.
The costs of these planned paths are then used to update the cost matrix \(M_c\) in \(Q\) (Lines 24-25).
Since \(M_c\) changes, the TA solution, the plan, and the cost should also be updated (Lines 26-27).

\begin{figure*}[t]
\centering
\includegraphics[width=1\textwidth]{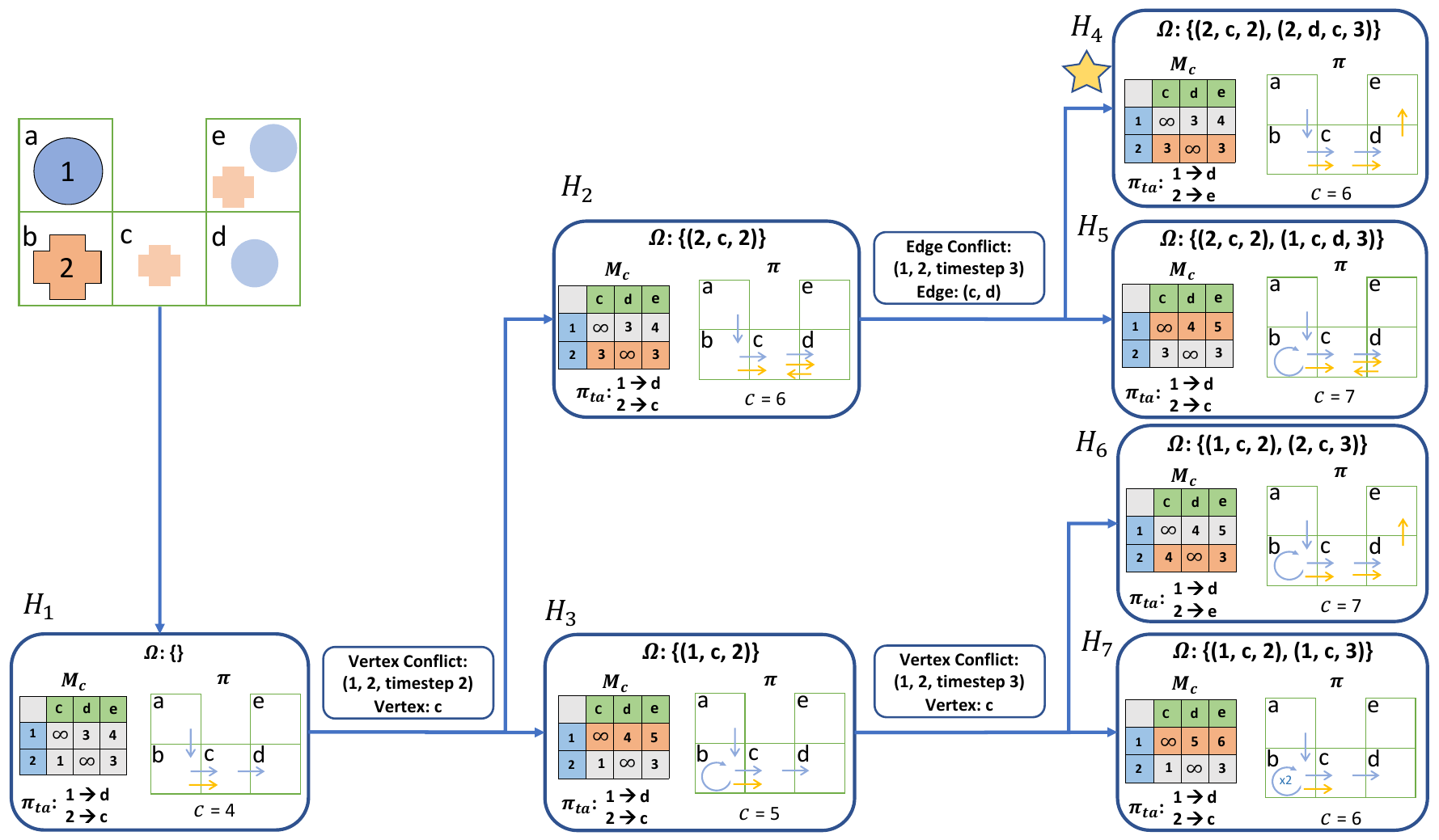} 
\caption{CT of ITA-CBS for a simple TAPF problem. The top left drawing shows a simple map with 5 cells ($a, b, c, d, e$) and 2 agents ($1, 2$). Start locations for agent $1$ and agent $2$ are $a$ and $b$, respectively. The target sets of agents $1$ and $2$ are $\{d, e\}$ and $\{c, e\}$, respectively. Each CT node \(H\) is represented by a blue rounded rectangle.}
\label{Example}
\end{figure*}

\subsection{Incremental Target Assignment}

In \Cref{alg:ITA-CBS1} Line 8, we use the Hungarian algorithm to get the TA solution for the root node.
The Hungarian algorithm solves bipartite graph matching optimally. A bipartite graph is a graph whose vertices can be decomposed into two disjoint sets such that no edges connect the vertices within the same set. In ITA-CBS, we form one vertex set with all $N$ agents and the other vertex set with all $M$ targets. We add an edge between an agent and a target if the corresponding entry in the cost matrix is finite.
The Hungarian algorithm assigns each vertex $v$ a value $l(v)$ such that $M_c[u][v] \leq l(u) + l(v)$ holds for every edge $(u,v)$.
An unweighted subgraph is then formed by including all vertices and edges satisfying the condition $M(u, v) = l(u) + l(v)$. It is proven that if the matching of this subgraph is a perfect matching, then this matching is an optimal matching in $M_c$~\cite{kuhn1955hungarian}. 
The Hungarian algorithm adjusts vertex values to achieve a perfect matching in this subgraph with a time complexity of $O(M^3)$.

While we can continue to use the Hungarian algorithm to get the TA solution in Line 26, running the Hungarian algorithm from scratch at every CT node is too costly for ITA-CBS.
In ITA-CBS, a child node contains only one new constraint on an agent compared to its parent node. Consequently, the cost matrix of the child node differs from that of the parent node only in the row pertaining to that particular agent.
Therefore, we employ the dynamic Hungarian algorithm~\cite{mills2007dynamic,amalia2021fast} to reuse the TA solution from the parent node. We unmatch the vertex pair corresponding to the particular agent and adjust the vertex value $l(i)$ for each affected vertex $i$, ensuring that $M(u, v) \leq l(u) + l(v)$ still holds. This dynamic Hungarian algorithm finds a new TA solution with a time complexity of $O(M^2)$, significantly faster than the Hungarian algorithm used by the root node of ITA-CBS (which is $O(M^3)$) and the K-best assignment used by CBS-TA (which is $O(N^2M^2)$). 


\subsection{Example}

Fig.\ref{Example} shows an example of our algorithm on a small map with 2 agents. 
To begin with, we generate the first node $H_1$ by calling the low-level search to get $M_c$, calling the  Hungarian algorithm to get $\pi_{ta}$ from $M_c$, and then obtaining \(\pi\) and $c$. Since there is no constraint in $\Omega$, agent 1 moves to $d$ in 3 timesteps, and agent 2 moves to $c$ in 1 timestep, leading to a vertex conflict at $c$ at timestep $2$. To resolve this conflict, two child nodes $H_2, H_3$ are created. With the new constraint added to $\Omega$ in each child node, we update $M_c$,$\pi_{ta}$, $\pi$, and $c$ corresondingly. Consequently, the node order in OPEN becomes \([H_3, H_2]\). 
Next, we expand $H_3$ and find a vertex conflict at $c$ at timestep $3$. We thus generate two nodes $H_6, H_7$. The updated OPEN is \([H_2, H_7, H_6]\). In $H_2$, we have an edge conflict along edge $(c,d)$ at timestep $3$. Upon addressing this conflict, OPEN becomes \([H_4, H_7, H_5, H_6]\). 
Within \(H_4\), $M_c$ exhibits two equal TA solutions: \(\{1 \rightarrow d, 1 \rightarrow c\}\) and \(\{1 \rightarrow d, 1 \rightarrow e\}\). Assume that we select the second TA solution. Finally, there is no conflict in \(H_4.\pi\), so we find am optimal solution with a flowtime of $6$.




\subsection{Properties of ITA-CBS}

This section shows that ITA-CBS is guaranteed to find an optimal TAPF solution if one exists.

\begin{lemma}
\label{lemma1}
The cost of each CT node is a lower bound on the flowtime of all solutions that satisfy the node's constraints.
\end{lemma}
\begin{proofsketch}
Consider a CT node \(H = (c, \Omega, \pi, \pi_{ta}, M_c)\). Let $\{p^i\}$ be an arbitrary solution that satisfies $\Omega$.  Since the entries of $M_c$ correspond to the costs of the shortest paths that satisfy $\Omega$, the cost of each path $p^i$ in $\{p^i\}$ is no smaller than the corresponding entry (i.e., the entry with the same start and target locations) of $M_c$. That is, the flowtime of $\{p^i\}$ is no smaller than the total assignment cost of the corresponding TA solution based on $M_c$. Since $\pi_{ta}$ is the optimal TA solution based on $M_c$, the flowtime of $\{p^i\}$ is no smaller than the total assignment cost of $\pi_{ta}$, which equals $c$. Therefore, the lemma holds.
\end{proofsketch}

\begin{lemma}
\label{lemma3}
Every solution that satisfies the constraints of a CT node must also satisfy the constraints of at least one of its child nodes.
\end{lemma}

\begin{proofsketch}
We prove by contradiction and assume that there is a solution $\{p^i\}$ that satisfies the constraints of a CT node ${H}$ but does not satisfy the constraints of either child node. Suppose the conflict chosen to resolve in $H$ is between agents $i$ and $j$ at vertex $v$ (or edge $e$) at timestep $t$. Since each child node has only one additional constraint compared to node $H$, we know that $\{p^i\}$ violates both additional constraints. That is, both path $p^i$ and path $p^j$ visit vertex $v$ (or edge $e$) at timestep $t$, which leads to a conflict and contradicts the assumption that $\{p^i\}$ is conflict-free. Therefore, the lemma holds.
\end{proofsketch}

\begin{lemma}
\label{lemma4}
During the search, every solution must satisfy the constraints of at least one CT node in the OPEN list.
\end{lemma}

\begin{proofsketch}
Since the root CT node has no constraints, all solutions satisfy the constraints of the root CT node. When we pop a CT node from the OPEN list, we insert its child nodes back into the OPEN list. According to \Cref{lemma4}, this lemma holds.
\end{proofsketch}

\begin{theorem}
\label{lemma5}
ITA-CBS guarantees to find an optimal TAPF solution if one exists.
\end{theorem}

\begin{proofsketch}
According to \Cref{lemma1,lemma4}, the minimum cost of the CT nodes in the OPEN list is a lower bound on the flowtime of all solutions. Thus, when ITA-CBS terminates, its returned solution is guaranteed to be optimal. 
\end{proofsketch}

\begin{figure*}[t]
\centering
\includegraphics[width=1\textwidth]{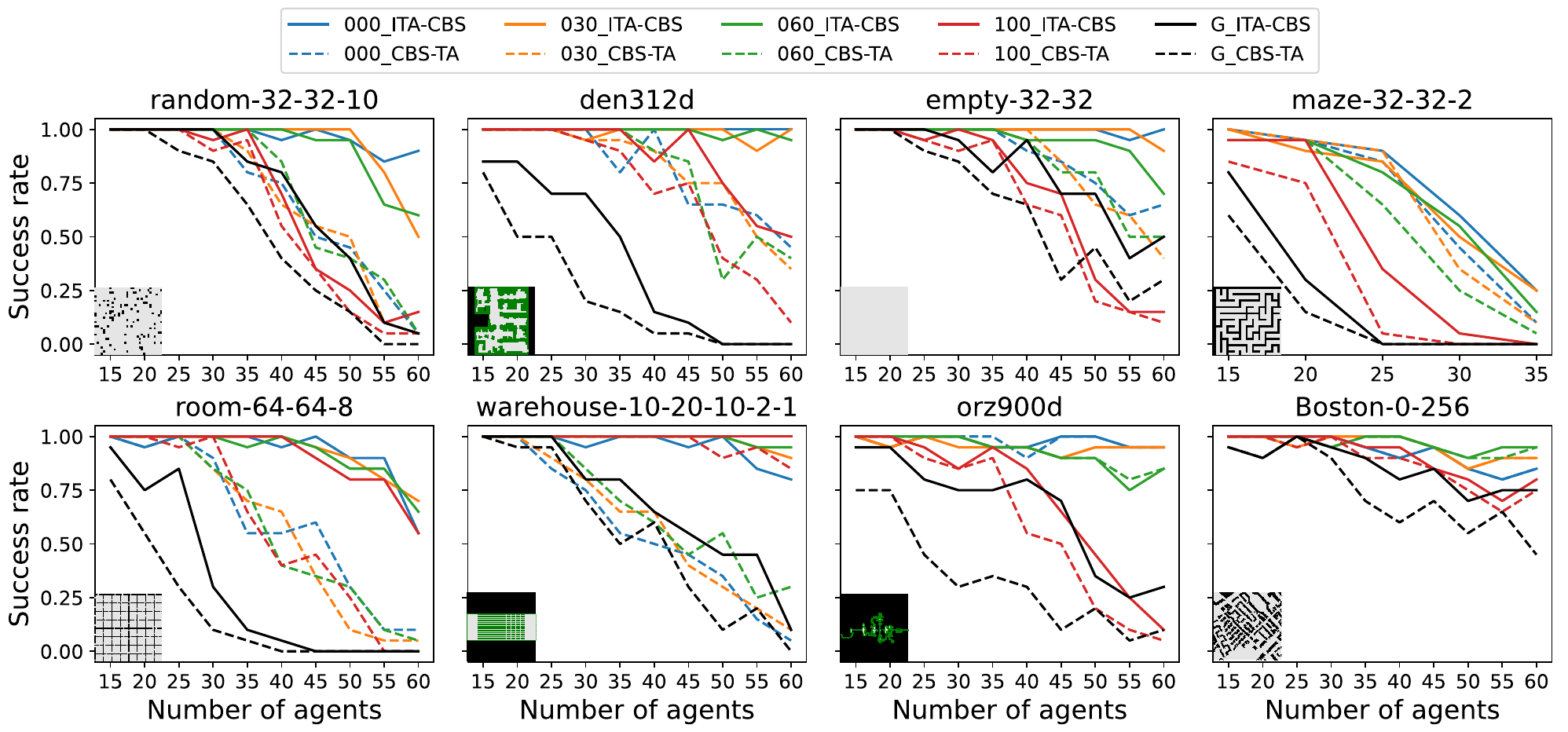} 
\caption{Success rates. In the legend, ``G\_'' indicates the group test results, while ``xxx\_'' indicates the common target test results. For instance, 000 indicates that there is no shared target, and 100 indicates that all agents share the same target set.}
\label{fig:test2}
\end{figure*}

\begin{figure}[t!]
\centering
\includegraphics[width=0.32\textwidth]{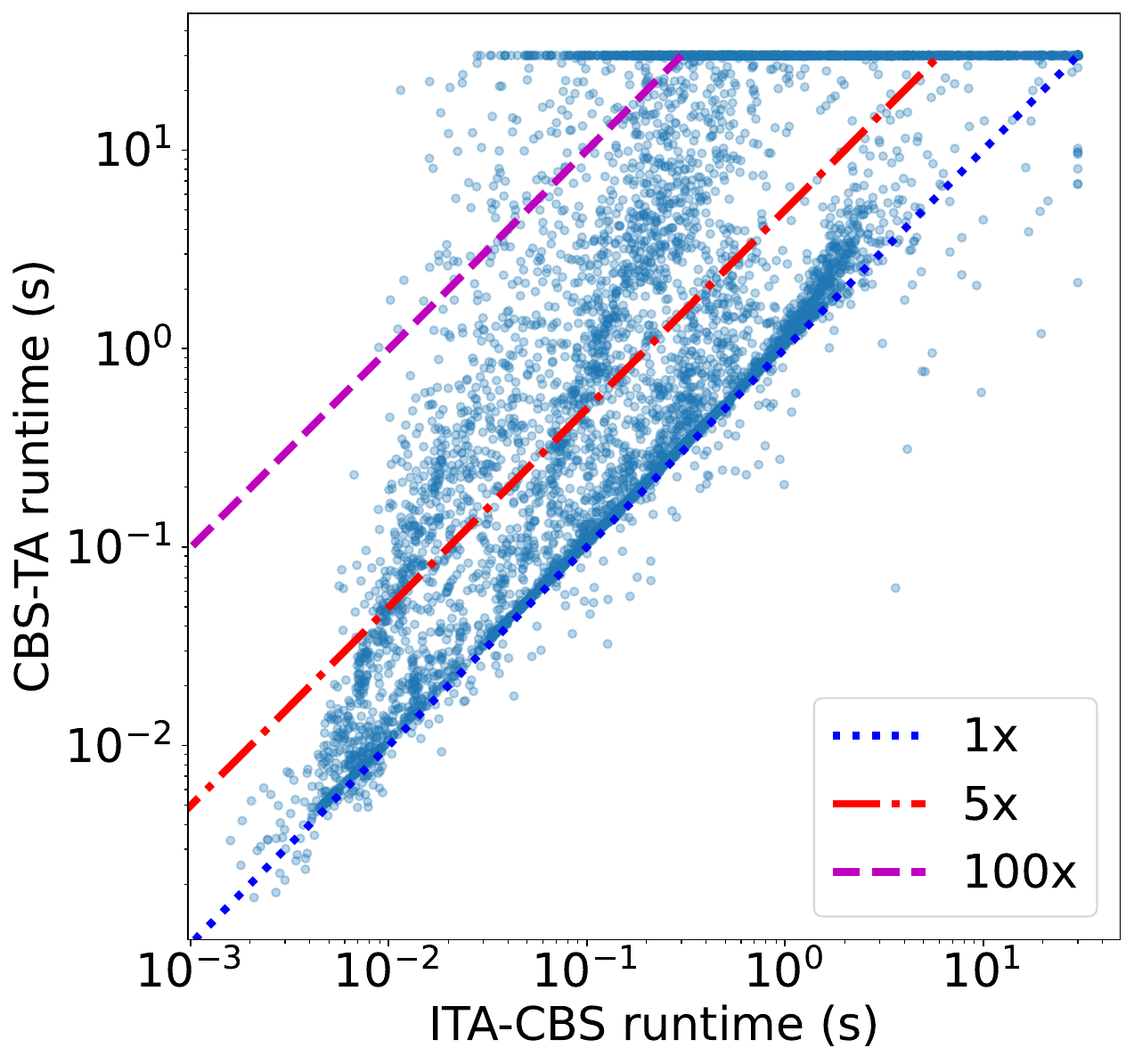} 
\caption{Runtime. 
We record the runtime as 30s for timeout test cases, so there is a line at the top of the figure. 
}
\label{fig:all_result}
\end{figure}

\begin{figure*}[htb]
\centering
     \begin{subfigure}[b]{0.32\textwidth}
         \centering
         \includegraphics[width=\textwidth]{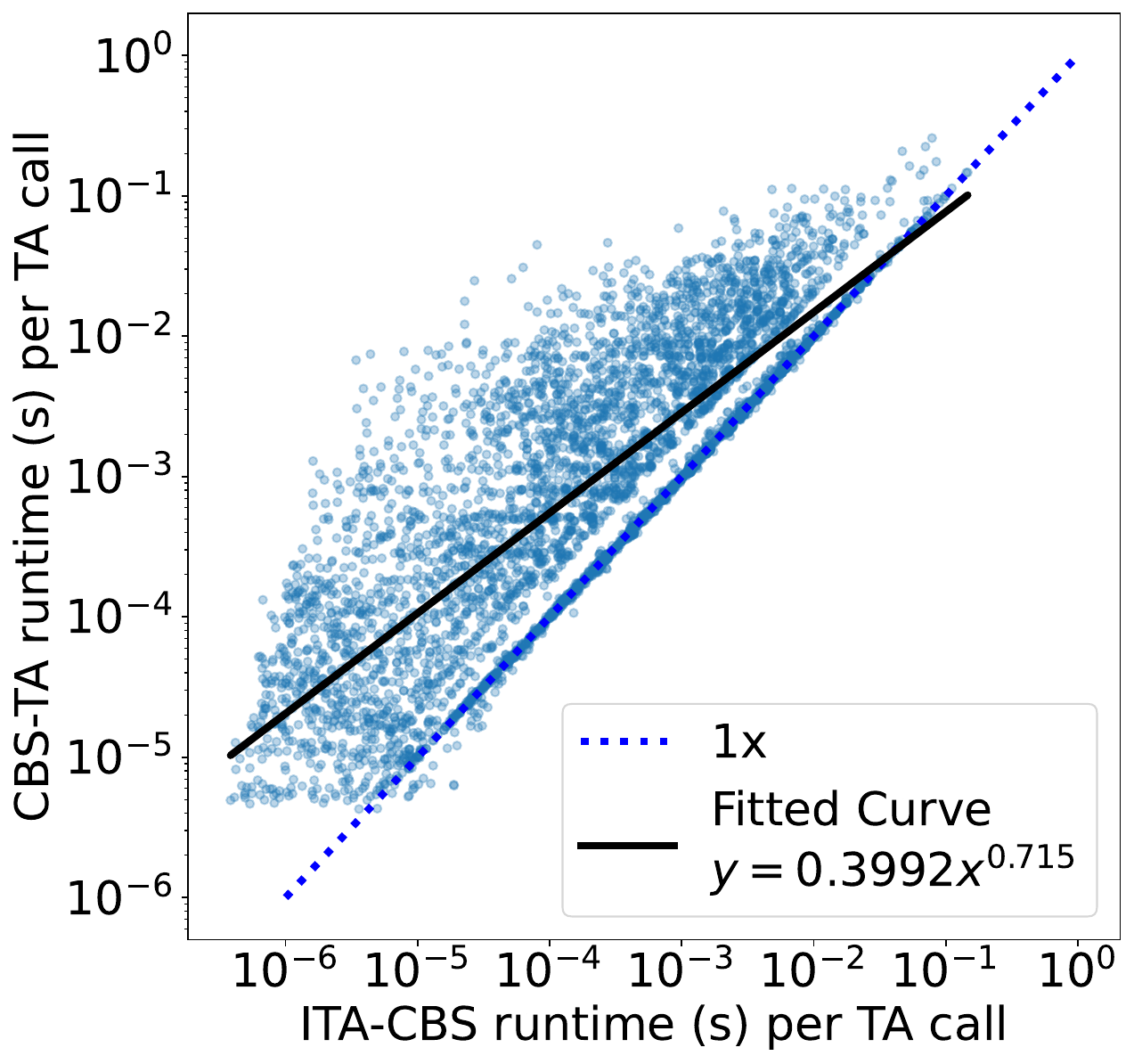}
         \label{fig:time_per_TA}
     \end{subfigure}
     \begin{subfigure}[b]{0.32\textwidth}
         \centering
         \includegraphics[width=\textwidth]{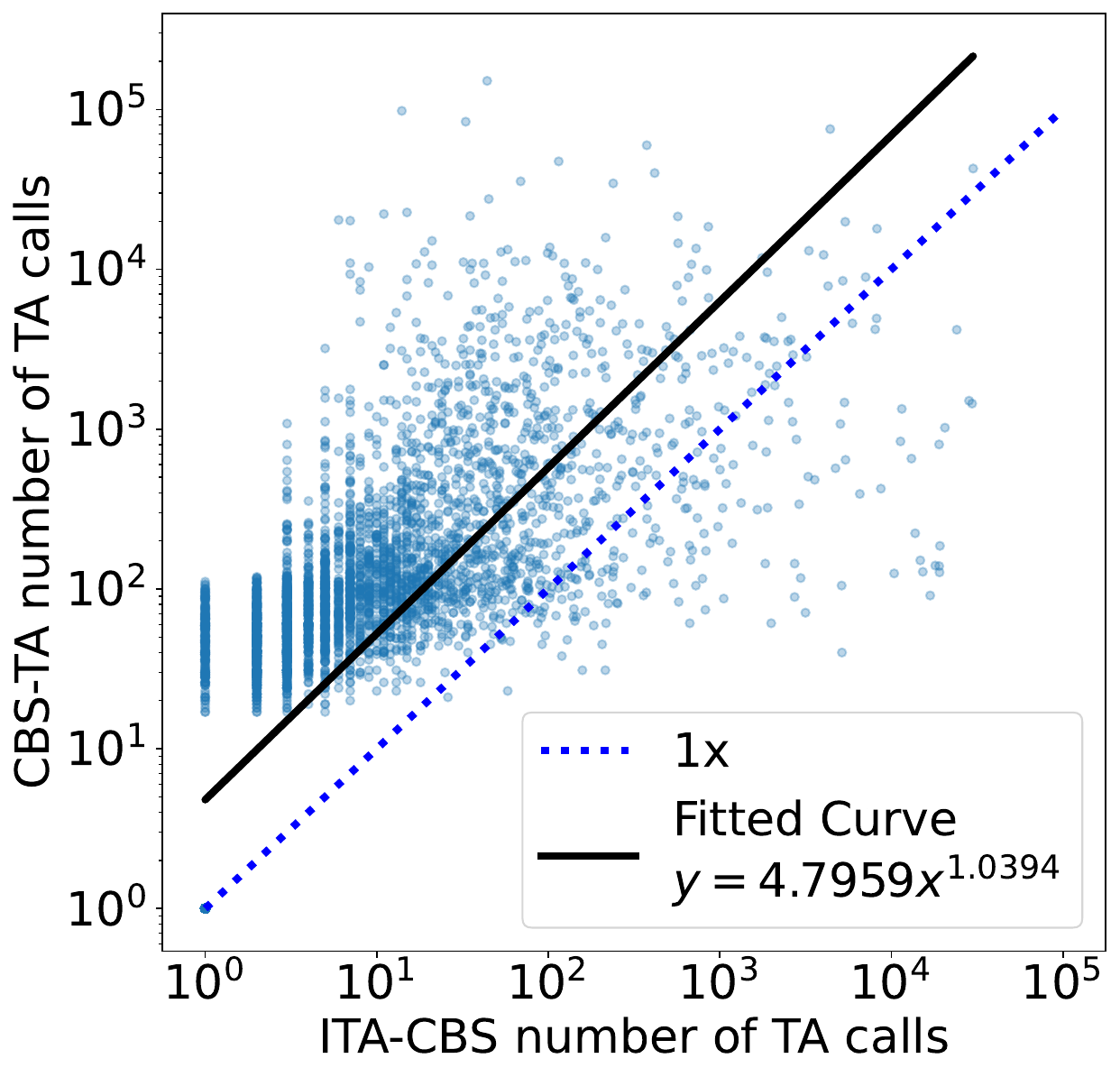}
         \label{fig:node_and_tree}
     \end{subfigure}
     \begin{subfigure}[b]{0.32\textwidth}
         \centering
         \includegraphics[width=\textwidth]{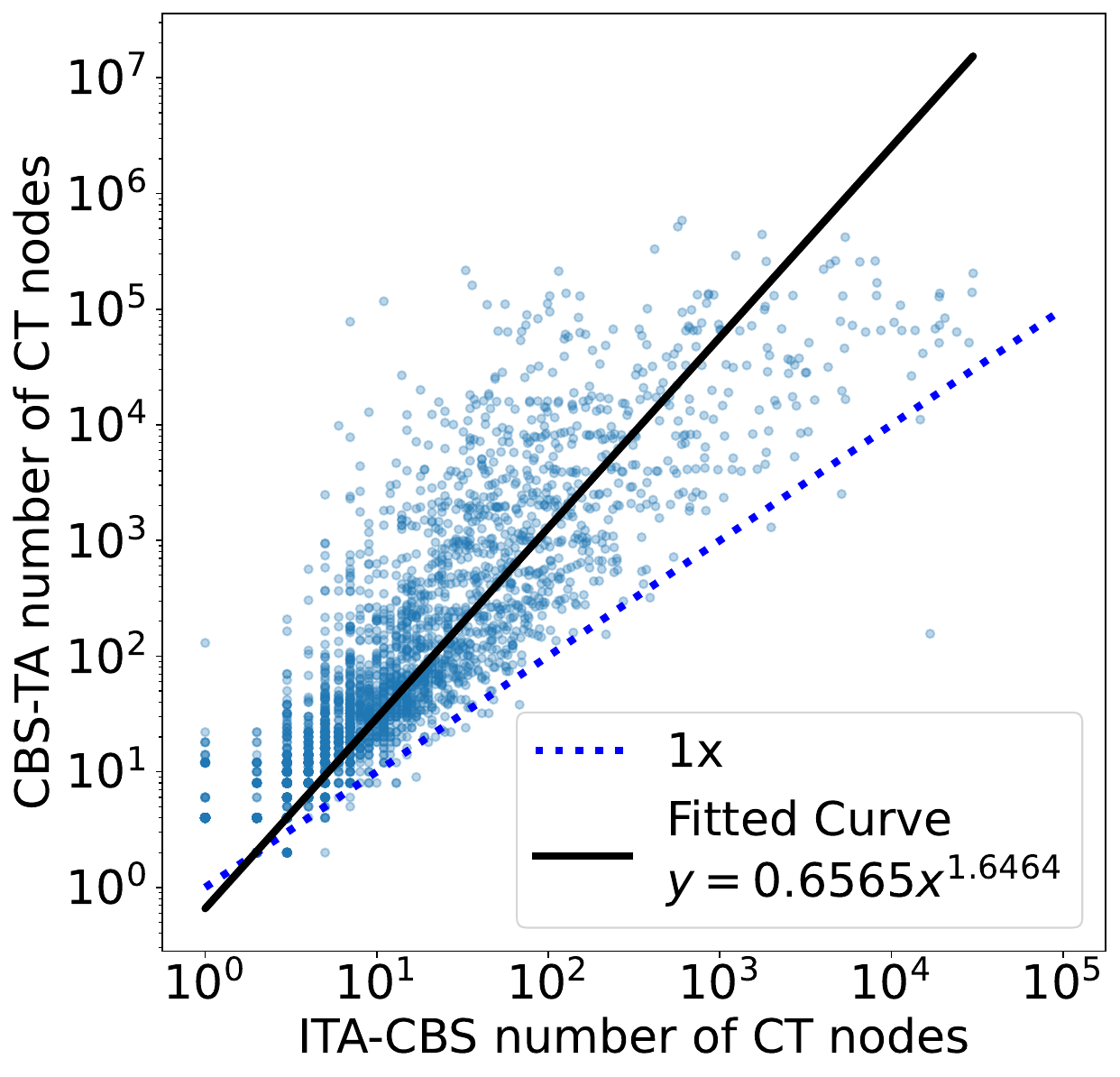}
         \label{fig:node_numbers}
     \end{subfigure}
    \caption{
    TA runtime and node expansions. 
    }
    \label{fig:test3_2}
\end{figure*}

\section{Experimental Results}

We compare the performance of ITA-CBS with CBS-TA since, to our best knowledge, CBS-TA is the only existing work that solves TAPF optimally for flowtime.
We implement both ITA-CBS and CBS-TA in C++ partially based on the existing CBS-TA implementation.\footnote{The CBS-TA source code is publicly available at \url{https://github.com/whoenig/libMultiRobotPlanning}.
Our code is available at \url{https://github.com/TachikakaMin/ITA-CBS2}.
Our CBS-TA implementation runs faster than the original one based on our tests.} 
All experiments were executed on a computer with Ubuntu 20.04.1, AMD Ryzen 3990X 64-Core Processor, 64G RAM with 2133 MHz.

We use 8 different maps, shown in Fig.\ref{fig:test2}, from the MAPF Benchmark sets~\cite{stern2019mapf}: (1) random-32-32-10 (32x32) and empty-32-32 (32x32) are open grids with and without random obstacles, (2) den312d (65x81) is from video game Dragon Age Origins, (3) maze-32-32-2 (32x32) is a maze-like grid, (4) room-64-64-8 (64x64), denoted by room is a room-like grid, (5) warehouse-10-20-10-2-1 (161x63) is inspired by real-world autonomous warehouse applications, and (6) orz900d (1491x656) and Boston-0-256 (256x256) are the first and second largest maps among all benchmark map files.

\subsection{Test Settings}

We design two types of test scenarios: (1) Group Test: We randomly divide agents into groups of size 5. Agents within the same group share a target set of size 5. Target sets from different groups do not contain any identical targets. (2) Common Target Test: For each map, every agent has a target set of the same size, which is 15, 40, 15, 15, 50, 80, 20, and 20 for maps random-32-32-10, den312d, empty-32-32, maze-32-32-2, room-64-64-8, warehouse-10-20-10-2-1, orz900d, and Boston-0-256, respectively.\footnote{The sizes of these target sets are determined by having the targets occupy all empty grid cells on the map under the 0\% scenario, except for large maps orz900d and Boston-0-256. On these large maps, the size of the target sets is limited to 20 to prevent both algorithms from timing out in any test case due to an excessive number of targets.} Each target set contains both targets shared among all agents and unique targets. We vary the ratio of shared targets in each target set from 0\%, 30\%, 60\%, to 100\%, resulting in four test scenarios. However, we ensure that each target set always includes at least one unique target to guarantee the existence of a solution.

For each test scenario, map, and number of agents, we generate 20 test cases with randomly selected start and target locations. An algorithm is considered to have failed for a given test case if it does not find an optimal solution within 30 seconds. The success rate is the percentage of the test cases where the algorithm succeeds out of the 20 test cases.

\subsection{Overall Performance}


Fig.\ref{fig:test2} shows the success rates. In the Group Test (black lines), ITA-CBS outperforms CBS-TA across all maps. In the Common Target Test, the success rates decrease for both algorithms as the ratio of the shared targets increases, but ITA-CBS still outperforms CBS-TA in almost all cases.

Fig.\ref{fig:all_result} shows the runtime. 
We have a total of 7,600 test cases, including 5,134 test cases solved by both algorithms, 1,191 test cases solved only by ITA-CBS, 9 test cases solved only by CBS-TA, and 1,266 test cases that both algorithms fail to solve. 
As shown, ITA-CBS is faster in 96.1\% test cases, 5 times faster in 38.7\% test cases, and 100 times faster in 5.6\% test cases than CBS-TA among the 6,334 test cases solved by at least one algorithm.

\subsection{Program Profile}

\begin{figure}[t!]
\centering
\includegraphics[width=0.4\textwidth]{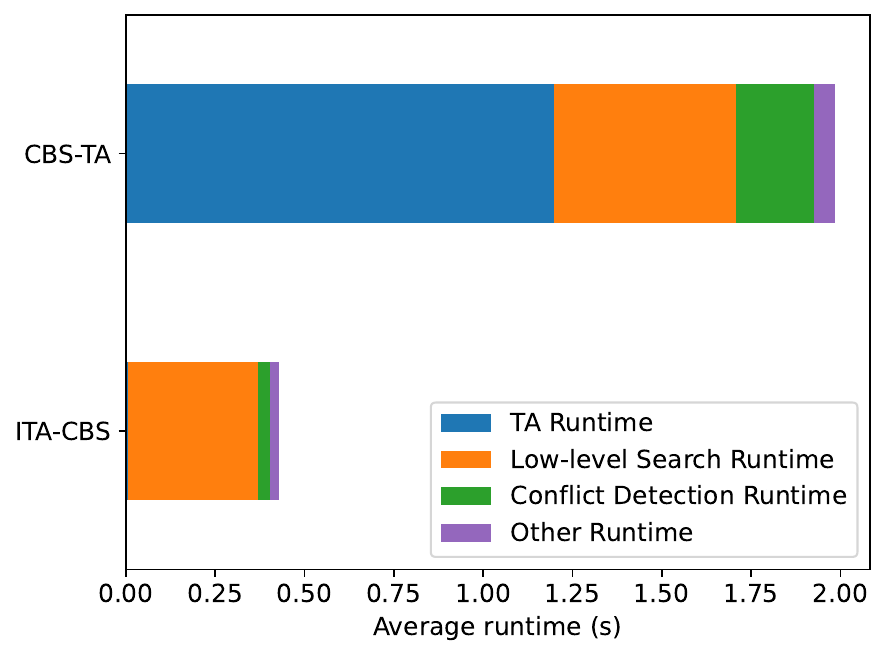}
\caption{Runtime breakdown. The runtime of TA, low-level search, conflict detection, and others for CBS-TA and ITA-CBS is \{$1.2s$, $0.51s$, $0.22s$, $0.058s$\} and \{$0.006s$, $0.36s$, $0.032s$, $0.027s$\}, respectively.}
\label{fig:test3}
\end{figure}

We compare the detailed performance of the two algorithms using the 5,134 test cases solved by both of them. 
In \Cref{fig:test3}, we show the average runtime for various parts of each algorithm and divide the algorithm runtime into 4 parts: 
TA runtime (\Cref{alg:ITA-CBS1} Lines 8, 26),
low-level search runtime (\Cref{alg:ITA-CBS1} Lines 3-7, 24-25),
conflict detection runtime (\Cref{alg:ITA-CBS1} Lines 14 and 17), and
others. 

ITA-CBS is faster than CBS-TA for all components, primarily due to its significantly reduced node expansions compared to CBS-TA. Notably, the TA runtime in ITA-CBS is 200 times smaller than that in CBS-TA, which is an interesting result since ITA-CBS calls TA algorithms at every CT node while CBS-TA calls TA algorithms only at roots.

To understand this result, Fig.\ref{fig:test3_2} (left) compares the average runtime per TA algorithm call for each test case. This shows that the TA algorithm in ITA-CBS (i.e., Dynamic Hungarian) is significantly faster than that in CBS-TA (i.e., K-best assignment). Fig.\ref{fig:test3_2} (middle) reveals another interesting result, where ITA-CBS requires fewer TA algorithm calls. This is primarily due to two factors: (1) ITA-CBS has significantly fewer node expansions than CBS-TA, as illustrated in \Cref{fig:test3_2} (right), and (2) CBS-TA often generates a substantial number of CTs; across 5,134 test cases, on average, 37.7\% of CT nodes generated by CBS-TA are roots.

\section{Conclusion}

This work develops a new algorithm called Incremental Target Assignment CBS (ITA-CBS) to solve the TAPF problem to optimality with flowtime.
ITA-CBS distinguishes itself from the prior leading algorithm, CBS-TA, in two key ways:
First, ITA-CBS constructs a single constraint tree throughout the search, leading to a reduction in CT nodes compared to CBS-TA. 
Second, ITA-CBS avoids solving the K-best assignment problem, and instead, it updates the target assignment in an incremental manner during the CBS-like search, which further reduces the computational effort.
We prove that ITA-CBS is optimal and show empirically that it runs significantly faster than CBS-TA.


\section{Acknowledgement}

This work has been funded in part by the Air Force Office of Scientific Research (AFOSR) under grants FA9550-18-1-0251 and FA9550-18-1-0097, the Army Research Laboratory (ARL) under grant W911NF-19-2-0146,  DARPA award HR001120C0036, and the CMU Manufacturing Futures Institute, made possible by the Richard King Mellon Foundation.

\bibliographystyle{IEEEtran} 
\bibliography{strings,IEEEabrv,myref}

\end{document}